\documentstyle[11pt,lingmacros]{article}

\makeatletter

\def\@citex[#1]#2{\if@filesw\immediate\write\@auxout{\string\citation{#2}}\fi
  \def\@citea{}\@cite{\@for\@citeb:=#2\do
    {\@citea\def\@citea{;\penalty\@m\ }\@ifundefined
       {b@\@citeb}{{\bf ?}\@warning
       {Citation `\@citeb' on page \thepage \space undefined}}%
{\csname b@\@citeb\endcsname}}}{#1}}
\let\@internalcite\cite
\def\cite{\def\citename##1{##1, }\@internalcite}
\def\shortcite{\def\citename##1{}\@internalcite}
\def\newcite{\leavevmode\def\citename##1{{##1} (}\@internalciteb}

\def\@citexb[#1]#2{\if@filesw\immediate\write\@auxout{\string\citation{#2}}\fi
  \def\@citea{}\@newcite{\@for\@citeb:=#2\do
    {\@citea\def\@citea{;\penalty\@m\ }\@ifundefined
       {b@\@citeb}{{\bf ?}\@warning
       {Citation `\@citeb' on page \thepage \space undefined}}%
\hbox{\csname b@\@citeb\endcsname}}}{#1}}
\def\@internalciteb{\@ifnextchar
[{\@tempswatrue\@citexb}{\@tempswafalse\@citexb[]}}

\def\@newcite#1#2{{#1\if@tempswa, #2\fi)}}

\def\@biblabel#1{\def\citename##1{##1}[#1]\hfill}

\def\@cite#1#2{({#1\if@tempswa , #2\fi})}

\def\thebibliography#1{\vskip\parskip%
\vskip\baselineskip%
\def\baselinestretch{1}%
\ifx\@currsize\normalsize\@normalsize\else\@currsize\fi%
\vskip-\parskip%
\vskip-\baselineskip%
\section*{References\@mkboth
 {References}{References}}\list
 {}{\setlength{\labelwidth}{0pt}\setlength{\leftmargin}{\parindent}
 \setlength{\itemindent}{-\parindent}}
 \def\newblock{\hskip .11em plus .33em minus -.07em}
 \sloppy\clubpenalty4000\widowpenalty4000
 \sfcode`\.=1000\relax}

\def\thesourcebibliography#1{\vskip\parskip%
\vskip\baselineskip%
\def\baselinestretch{1}%
\ifx\@currsize\normalsize\@normalsize\else\@currsize\fi%
\vskip-\parskip%
\vskip-\baselineskip%
\section*{Sources of Attested Examples\@mkboth
 {Sources of Attested Examples}{Sources of Attested Examples}}\list
 {}{\setlength{\labelwidth}{0pt}\setlength{\leftmargin}{\parindent}
 \setlength{\itemindent}{-\parindent}}
 \def\newblock{\hskip .11em plus .33em minus -.07em}
 \sloppy\clubpenalty4000\widowpenalty4000
 \sfcode`\.=1000\relax}

\def\@lbibitem[#1]#2{\item[]\if@filesw
      { \def\protect##1{\string ##1\space}\immediate
        \write\@auxout{\string\bibcite{#2}{#1}}\fi\ignorespaces}}

\def\@bibitem#1{\item\if@filesw \immediate\write\@auxout
       {\string\bibcite{#1}{\the\c@enumi}}\fi\ignorespaces}

\makeatother

\author{Andrew Kehler \\ Harvard University \\
Aiken Computation Laboratory
\\ 33 Oxford Street \\ Cambridge, MA 02138 \\ kehler@das.harvard.edu}
\date{}
\title{\vspace{-.3in} \LARGE\bf TEMPORAL RELATIONS:  \\ REFERENCE OR
DISCOURSE COHERENCE?
\\ \vspace{.12in} \small {\it (To appear in the Proceedings of ACL-94, Student
Session)}}

\begin{document}

\maketitle

\bibliographystyle{acl}

\begin{abstract}
The temporal relations that hold between events described by
successive utterances are often left implicit or underspecified.  We
address the role of two phenomena with respect to the recovery of
these relations: (1) the referential properties of tense, and (2) the
role of temporal constraints imposed by coherence relations.  We
account for several facets of the identification of temporal
relations through an integration of these.
\end{abstract}

\section{Introduction}

Tense interpretation has received much attention in linguistics
\cite[inter alia]{Partee:84,Hinrichs:86,Nerbonne:86} and nat\-u\-ral lang\-uage
processing
\cite[inter alia]{Webber:88,KamPasPoe:93,LasAsh:93}.
Several researchers
\cite{Partee:84,Hinrichs:86,Nerbonne:86,Webber:88} have
sought to explain the temporal relations induced by tense by treating
it as anaphoric, drawing on Reichenbach's separation between event,
speech, and reference times \cite{Reichenbach:47}.  Specifically, to
account for the forward progression of time induced by
successive simple past tenses in a narrative, they treat the simple
past as referring to a time evoked by a previous past tense.  For
instance, in Hinrichs's \shortcite{Hinrichs:86} proposal,
accomplishments and achievements\footnote{We will limit the scope of
this paper by restricting the discussion to accomplishments and
achievements.} introduce a new reference point that
is temporally ordered after the time of the event itself, ``ensuring
that two consecutive accomplishments or achievements in a discourse
are always ordered in a temporal sequence.''
On the other hand, Lascarides and Asher \shortcite{LasAsh:93} take the
view that temporal relations are resolved purely as a by-product of
reasoning about coherence relations holding between utterances, and in
doing so, argue that treating simple and complex tenses as
anaphoric is unnecessary.  This approach parallels the treatment of
pronoun resolution espoused by Hobbs \shortcite{Hobbs:78a}, in which
pronouns are modeled as free variables that are bound as a by-product
of coherence resolution.  The Temporal Centering framework
\cite{KamPasPoe:93} integrates aspects of both
approaches, but patterns with the first in treating tense as
anaphoric.

We argue that aspects of both analyses are necessary to account for
the recovery of temporal relations.  To demonstrate our approach we
will address the following examples; passages (\ref{bucket}a-b) are
taken from Lascarides and Asher \shortcite{LasAsh:93}:

\eenumsentence{ \label{bucket}
\item Max slipped.  He spilt a bucket of water.

\item Max slipped.  He had spilt a bucket of water.

\item Max slipped because he spilt a bucket of water.

\item Max slipped because he had spilt a bucket of water.
}
Passage (\ref{bucket}a) is understood as a {\it narrative}, indicating that
the spilling was subsequent to the slipping.  Passages
(\ref{bucket}b-d) are instead understood as the second clause {\it
explaining} the first, indicating that the reverse temporal ordering
holds.  We address two related questions; the first arises from
treating the simple past as anaphoric.  Specifically, if a treatment
such as Hinrichs's is used to explain the forward progression of time
in example (\ref{bucket}a), then it must be explained why sentence
(\ref{bucket}c) is as felicitous as sentence (\ref{bucket}d).  That
is, one would predict a clash of temporal relations for sentence
(\ref{bucket}c), since the simple pasts induce the forward progression
of time but the conjunction indicates the reverse temporal ordering.
The second question arises from assuming that all temporal relations
are recovered solely from reasoning with coherence relations.
Specifically, because the use of the simple past in passage
(\ref{bucket}c) is as felicitous as the past
perfect in passage (\ref{bucket}d) under the {\it explanation}
interpretation (in these cases indicated explicitly by {\it because}),
then it must be explained why passage (\ref{bucket}a) is not
understood as an {\it explanation} as is passage (\ref{bucket}b), where in
each case the relationship needs to be inferred.  We present our
analysis in the next section, and account for these facts in
Section~3.

\section{The Account}

We postulate rules characterizing the referential nature of tense and
the role of discourse relations in further constraining the temporal
relations between clauses.  The rules governing tense are:

\begin{enumerate}
\item Main verb tenses are indefinitely referential, creating a new temporal
entity under constraints imposed by its type (i.e., past,
present, or future) in relation to a {\it discourse reference
time}\footnote{This term is borrowed from Kameyama et al.
\shortcite{KamPasPoe:93}.} $t_{R}$.  For instance, a main verb
past tense introduces a new temporal entity $t$ under the constraint
{\it prior-to(t,t$_{R}$)}.  For simple tenses $t_{R}$ is the speech
time, and therefore simple tenses are not anaphoric.

\item Tensed auxiliaries in complex tenses are ana\-phor\-ic, ident\-ifying
$t_{R}$ as a
previously existing temporal entity.  The indefinite main verb tense
is then ordered with respect to this $t_{R}$.
\end{enumerate}
The tenses used may not completely specify the implicit temporal
relations between the described events.  We claim that these relations
may be further refined by constraints imposed by the coherence
relation operative between clauses.  We describe three coherence
relations relevant to the examples in this paper and give
temporal constraints for them.\footnote{We assume here that the two
clauses in question are related directly by a coherence relation.
This may not be the case; for instance the use of a past perfect may
signal the start of an embedded discourse segment, as in Webber's
flower shop example \cite{Webber:88,KamPasPoe:93}.  How this account
is to be extended to address coherence at the discourse segment level
is the subject of future work.}

\begin{description}

\item[Narration:]

The {\it Narration} relation is characterized by a series of events
displaying forward movement of time, such as in passage
(\ref{bucket}a). As did Lascarides and Asher \shortcite{LasAsh:93}, we
capture this ordering as a constraint imposed by the Narration
coherence relation itself:\footnote{The {\it Cause-Effect} relation
also has this ordering constraint.}

\enumsentence{If $Narration(A,B)$ then $t_{A} < t_{B}$}

\item[Parallel:]

The {\it Parallel} relation relates utterances that share a common topic.  This
relation does not impose constraints on the temporal relations between the
events beyond those provided by the tenses themselves.  For instance,
if passage (\ref{bucket}a) was uttered in response to the question
{\it What bad things happened to Max today?} (inducing a Parallel
relation instead of Narration), a temporal ordering among the
sentences is no longer implied.

\item[Explanation:]

The {\it Explanation} relation denotes a cause-effect relationship with
reversed clause ordering, as in sentences (\ref{bucket}b-d).
Therefore, the second event is constrained to preceding the
first:

\enumsentence{If {\it Explanation(A,B)} then $t_{B} < t_{A}$}

\end{description}

To summarize the analysis, we claim that tense operates as indefinite
reference with respect to a possibly anaphorically-resolved discourse
reference time.  The temporal relations specified may be further
refined as a by-product of establishing the coherence relationship
extant between clauses, {\it Narration} being but one such relation.

\section{Examples}

We now analyze the examples presented in Section 1, repeated below,
using this approach:

\eenumsentence{ \label{bucket2}
\item Max slipped.  He spilt a bucket of water.

\item Max slipped.  He had spilt a bucket of water.

\item Max slipped because he spilt a bucket of water.

\item Max slipped because he had spilt a bucket of water.
}

The implicit ordering on the times indefinitely evoked by the simple
pasts in passage (\ref{bucket2}a) results solely from understanding it
as a Narration.  In passage (\ref{bucket2}b), the auxiliary {\it had}
refers to the event time of the slipping, and thus the past tense on
{\it spill} creates a temporal entity constrained to precede that
time.  This necessitates a coherence relation that is consistent with this
temporal order, in this case, Explanation.  In passage
(\ref{bucket2}c), the times evoked by the simple pasts are further
ordered by the Explanation relation indicated by {\it because},
resulting in the backward progression of time.  In passage
(\ref{bucket2}d), both the tense and the coherence relation order the
times in backward progression.

Restating the first problem noted in Section 1, if treating the simple
past as anaphoric is used to account for the forward progression of
time in passage (\ref{bucket2}a), then one would expect the existence
of the Explanation relation in passage (\ref{bucket2}c) to cause a
temporal clash, where in fact passage (\ref{bucket2}c) is perfectly
felicitous.  No clash of temporal relations is predicted by our
account, because the use of the simple pasts do {\it not} in
themselves imply a specific ordering between them.  The Narration
relation orders the times in forward progression in passage
(\ref{bucket2}a) and the Explanation relation orders them in backward
progression in passage (\ref{bucket2}c). The Parallel relation would
specify no ordering (see the potential context for passage
(\ref{bucket2}a) given in Section 2).

Restating the second problem noted in Section 1, if temporal relations
can be recovered solely from reasoning with coherence relations, and
the use of the simple past in passage (\ref{bucket2}c) is as
felicitous as the past perfect in passage
(\ref{bucket2}d) under the Explanation interpretation, then one asks
why passage (\ref{bucket2}a) is not understood as an Explanation as is
passage (\ref{bucket2}b), where in each case the relationship needs to
be inferred.  We hypothesize that hearers assume that speakers are
engaging in Narration in absence of a specific cue to the contrary.
The use of the past perfect (as in passage (\ref{bucket2}b)) is one
such cue since it implies reversed temporal ordering; the use of an
explicit conjunction indicating a coherence relation other than
Narration (as in passages (\ref{bucket2}c-d)) is another such cue.
While passage (\ref{bucket2}a) could be understood as an Explanation
on semantic grounds, the hearer assumes Narration since no other
relation is cued.

We see several advantages of this approach over that of Lascarides and
Asher \shortcite[henceforth L\&A]{LasAsh:93}.  First, L\&A note the
incoherence of example (\ref{coffee})
\enumsentence{? Max poured a cup of coffee.  He had entered the room.
\label{coffee} }
in arguing that the past perfect should not be treated as anaphoric:
\enumsentence{Theories that analyse the distinction between the
simple past and pluperfect purely in terms of different relations
between reference times and event times, rather than in terms of
event-connections, fail to explain why [(\ref{bucket2}b)] is acceptable
but [(\ref{coffee})] is awkward.  \cite[pg. 470]{LasAsh:93}}
Example (\ref{coffee}) indeed shows that coherence relations need to be
utilized to account for temporal relations, but it does not bear on
the issue of whether the past perfect is anaphoric.  The incoherence
of example (\ref{coffee}) is predicted by both their and our accounts
by virtue of the fact that there is no coherence relation that
corresponds to Narration with reverse temporal ordering.\footnote{For
instance, in the same way that Explanation corresponds to Cause-Effect
with reverse temporal ordering.} In addressing this example,
L\&A specify a special rule (the {\it Connections When
Changing Tense (CCT)} Law) that stipulates that a sentence containing
the simple past followed by a sentence containing the past perfect can
be related only by a subset of the otherwise possible coherence relations.
However, this subset contains just those relations that are predicted
to be possible by accounts treating the past perfect as anaphoric;
they are the ones that do not constrain the temporal order of the
events against displaying backward progression of time.  Therefore, we
see no advantages to adopting their rule; furthermore,
they do not comment on what other laws have to be stipulated to
account for the facts concerning other possible tense combinations.

Second, to explain why the Explanation relation can be inferred for
passage (\ref{bucket2}b) but not for passage (\ref{bucket2}a),
L\&A stipulate that their causal {\it Slipping Law}
(stating that spilling can cause slipping) requires that the CCT Law
be satisfied.  This constraint is imposed only to require that the
second clause contain the past perfect instead of the simple past.
However, this does not explain why the use of the simple past is
perfectly coherent when the Explanation relationship is indicated
overtly as it is in sentence (\ref{bucket2}c), nor do they adequately
explain why CCT must be satisfied for this causal law and not for
those supporting similar examples for which they successfully
infer an unsignaled Explanation relation (see discussion of example
(2), pg.~463).

Third, the L\&A account does not explain why the
past perfect cannot stand alone nor discourses generally be opened
with it; consider stating sentence (\ref{pp-alone}) in isolation:

\enumsentence{Max had spilt a bucket of water. \label{pp-alone}}
Intuitively, such usage is infelicitous because of a dependency on a
contextually salient time which has not been previously introduced.
This is not captured by the L\&A account because
sentences containing the past perfect are treated as sententially
equivalent to those containing the simple past.  On the other hand,
sentences in the simple past are perfectly felicitous in standing
alone or opening a discourse, introducing an asymmetry in accounts
treating the simple past as anaphoric to a previously evoked time.
All of these facts are explained by the account given here.

\section{Conclusion}

We have given an account of temporal relations whereby (1) tense is
resolved indefinitely with respect to a possibly
anaphorically-resolved discourse reference time, and (2) the resultant
temporal relations may be further refined by constraints that
coherence relations impose.  This work is being expanded to
address issues pertaining to discourse structure and inter-segment
coherence.

\section*{Acknowledgments}

This work was supported in part by National Science Foundation Grant
IRI-9009018, National Science Foundation Grant IRI-9350192, and a
grant from the Xerox Corporation.  I would like to thank Stuart
Shieber and Barbara Grosz for valuable discussions and comments on
earlier drafts.


\begin{thebibliography}{}

\bibitem[\protect\citename{Hinrichs}1986]{Hinrichs:86}
Erhard Hinrichs.
\newblock 1986.
\newblock Temporal anaphora in discourses of english.
\newblock {\em Linguistics and Philosophy}, 9:63--82.

\bibitem[\protect\citename{Hobbs}1979]{Hobbs:78a}
Jerry Hobbs.
\newblock 1979.
\newblock Coherence and coreference.
\newblock {\em Cognitive Science}, 3:67--90.

\bibitem[\protect\citename{Kameyama \bgroup et al.\egroup }1993]{KamPasPoe:93}
Megumi Kameyama, Rebecca Passoneau, and Massimo Poesio.
\newblock 1993.
\newblock Temporal centering.
\newblock In {\em Proceedings of the 31st Conference of the Association for
  Computational Linguistics (ACL-93)}, pages 70--77, Columbus, Ohio, June.

\bibitem[\protect\citename{Lascarides and Asher}1993]{LasAsh:93}
Alex Lascarides and Nicolas Asher.
\newblock 1993.
\newblock Temporal interpretation, discourse relations, and common sense
  entailment.
\newblock {\em Linguistics and Philosophy}, 16(5):437--493.

\bibitem[\protect\citename{Nerbonne}1986]{Nerbonne:86}
John Nerbonne.
\newblock 1986.
\newblock Reference time and time in narration.
\newblock {\em Linguistics and Philosophy}, 9:83--95.

\bibitem[\protect\citename{Partee}1984]{Partee:84}
Barbara Partee.
\newblock 1984.
\newblock Nominal and temporal anaphora.
\newblock {\em Linguistics and Philosophy}, 7:243--286.

\bibitem[\protect\citename{Reichenbach}1947]{Reichenbach:47}
Hans Reichenbach.
\newblock 1947.
\newblock {\em Elements of Symbolic Logic}.
\newblock Macmillan, New York.

\bibitem[\protect\citename{Webber}1988]{Webber:88}
Bonnie~Lynn Webber.
\newblock 1988.
\newblock Tense as discourse anaphor.
\newblock {\em Computational Linguistics}, 14(2):61--73.

\end{thebibliography}
\end{document}